\pdfoutput=1

\documentclass[11pt]{article}

\usepackage[final]{acl}

\usepackage{times}
\usepackage{latexsym}

\usepackage[T1]{fontenc}

\usepackage[utf8]{inputenc}

\usepackage{microtype}

\usepackage{inconsolata}

\usepackage{graphicx}

\usepackage{blindtext}
\usepackage{booktabs}
\usepackage{caption}
\usepackage{subcaption}
\usepackage{csquotes}

\title{Resilience through Scene Context in Visual Referring Expression Generation}

\author{Simeon Junker \and Sina Zarrieß \\
  Computational Linguistics, Department of Linguistics\\
  Bielefeld University, Germany\\
  \texttt{\{simeon.junker,sina.zarriess\}@uni-bielefeld.de}\\
  }

\begin{document}
\maketitle
\begin{abstract}
Scene context is well known to facilitate humans' perception of visible objects.
In this paper, we investigate the role of context in Referring Expression Generation (REG) for objects in images, where existing research has often focused on distractor contexts that exert pressure on the generator.
We take a new perspective on scene context in REG and hypothesize that contextual information can be conceived of as a resource that makes REG models more resilient and facilitates the generation of object descriptions, and object types in particular. We train and test Transformer-based REG models with target representations that have been artificially obscured with noise to varying degrees.
We evaluate how properties of the models' visual context affect their processing and performance.
Our results show that even simple scene contexts make models surprisingly resilient to perturbations, to the extent that they can identify referent types even when visual information about the target is completely missing.\footnote{Code, models and data for this project are available at: \href{https://github.com/clause-bielefeld/REG-Scene-Context}{https://github.com/clause-bielefeld/REG-Scene-Context}}
\end{abstract}

\section{Introduction}

Objects do not appear randomly in the world that surrounds us, but they occur in predictable spatial, semantic, or functional configurations and relations to their environment. 
Research on human perception shows that we ``see the world in scenes'' \citep{Bar2004}, and that prior experience and knowledge of the world helps us to efficiently process visual stimuli.
Even with an extremely short glimpse at an image, humans remember essential semantic aspects of the scene and object arrangement \citep{Oliva2006}.
This rapid scene understanding allows us to handle the complexity of the visual world and to recognize objects in context, e.g., when they are not fully visible \citep{Vo2021}.

Today's systems for Vision and Language (V\&L) commonly process visual inputs that represent \enquote{real-world} scenes (e.g. \citealt{Lin2014,Antol2015,Krishna2016,Das2017}) which, to some extent, exhibit the regularities that human perception is known to be exploiting. Yet, it is not clear \textit{how} current V\&L systems process context and whether they rely on strategies of scene understanding similar to humans. 
In this paper, we aim to investigate this question for Referring Expression Generation (REG, \citealt{Dale1995,Mao2016}), a controlled set-up that is well established in NLG research, by testing how scene context supports 
reference generation for objects that are difficult to recognize.

\begin{figure}
    \centering
    \includegraphics[width=.8\columnwidth]{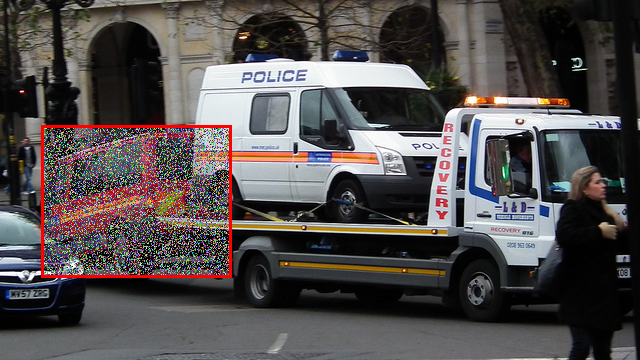}
    \addtolength{\tabcolsep}{-4pt}    
    \footnotesize
        \begin{tabular}{lllr}
                  & TRF$_{tgt}$ & red van (A)\\
        noise 0.0 & TRF$_{vis}$ & red truck (A)\\
                  & TRF$_{sym}$ & red truck (A)\\
        \midrule
                  & TRF$_{tgt}$ & left elephant (F)\\
        noise 1.0 & TRF$_{vis}$ & white truck (A)\\
                  & TRF$_{sym}$ & car on left (A)\\
    \end{tabular}
    \caption{
    Example from RefCOCO (displayed with noise level $0.5$) with generated expressions and human judgments. Visual or symbolic scene context 
    allows to identify even fully occluded targets (noise $1.0$).
    }
    \label{fig:example}
\end{figure}

Whereas classical REG algorithms mostly build on pre-defined symbolic representations \citep{Krahmer2012}, neural generation models in \textit{visual} REG have to extract object properties from low-level visual representations (i.e., photographs) of the target and its context \citep{Schuez2023}.
This even applies to properties as fundamental as the \textit{type} of an object, i.e. how it is \textit{named} in the expression. Under ideal conditions, determining a referent's type and properties can be regarded as a relatively simple task, but it becomes non-trivial in the presence of imperfect visual information, occlusion or noise.
Here, in light of previous findings on human scene understanding (cf. Section \ref{sec:background}), scene context can be expected to be of great support. However, to date, little is known as to how processes of scene understanding and object type identification interact in REG.

In this work, we hypothesize that visual scene context makes REG models more \textit{resilient}, i.e., it allows them to recalibrate predictions that were based on imperfect target representations.
To test this, we use a novel and highly controllable experimental setup for REG:
we train and test different Transformer-based model architectures with target representations that have been artificially obscured with varying degrees of noise (cf. Figure \ref{fig:example}), simulating scenarios that are common in the real world but insufficiently represented in current REG datasets. We provide the models with different context representations and compare their performance on common quality metrics and a focused human evaluation of their ability to determine referent types. 
Our results show that context makes models surprisingly resilient to perturbations in target representations, to the extent that they can identify referent types even when information about the objects themselves is completely missing. 
We believe that these results open up new perspectives on how information about the structure and content of surrounding scenes facilitate the description of objects in REG and related tasks.

\section{Background}
\label{sec:background}

\paragraph{Human scene understanding}

Research on human vision and perception emphasizes the fact that scenes are not mere collections of objects \citep{Vo2021}. When humans \textit{view} a scene, they do not simply recognize the objects in it, but \textit{understand} it as a coherent whole. \citet{Oliva2006} observe that humans perceive the so-called gist of a scene rapidly and even when local information is missing (e.g. blurred). Other experiments indicate that contextual information can facilitate the recognition of visible objects across different tasks \citep{Oliva2007,Divvala2009,Galleguillos2010,Parikh2012}, and that incongruent context can also be misleading \citep{Zhang2020a, Gupta2022} demonstrating that the human vision exploits learned knowledge about regularities of the visual word for visual processing \citep{Biederman1972,Bar2004,Greene2013,Pereira2014,Sadeghi2015,Vo2021}.

\paragraph{Scenes, objects, and image captioning}
Much research on V\&L is concerned with modeling the generation and understanding of image descriptions, e.g. in image captioning (\citealt{Xu2015,Anderson2018,Cornia2020}, among many others). Yet,  many captioning tasks focus on rather object-centric descriptions that mention objects and their spatial relationships \citep{Cafagna2021}. A common representation of scene context in image captioning is scene graphs \citep{yang-etal-2023-transforming}, which are usually modeled via spatial relations between bounding boxes of objects.
\citealt{cafagna-etal-2023-hl}  propose a new task and dataset that foregrounds scene-level instead of object-centric descriptions.
Another perspective on scene knowledge in captioning models is coming from work that focuses on probing them with perturbed or systematically varied images: \citet{yin-ordonez-2017-obj2text} find that captioning with extremely reduced inputs of labeled object layouts performs surprisingly well. Related to this, \citet{nikolaus-etal-2019-compositional} show that image captioning models often rely on regularities in object occurrences, to the extent that they fail to generalize to new combinations of objects. Their solution is to generate unseen combinations and challenge models on these. Our goal in this work is complementary: we aim to understand how exactly generation models may be able to leverage regular scene knowledge and patterns of object co-occurrence, and how this may facilitate the handling of imperfect visual information.

\paragraph{REG and scene context}

REG is concerned with the generation of descriptions that distinguish a particular object in a given visual context, cf. \citealt{Krahmer2012}.
In past years, REG research has largely transitioned from symbolic settings to \textit{visual REG}, focusing on referring expressions for objects in photographs \citep{Kazemzadeh2014, Mao2016}.
Recent models usually build on image captioning models but are adapted to generate more pragmatically informative expressions, using e.g. training objectives \citep{Mao2016}, comprehension modules \citep{Luo2017}, reinforcement agents \citep{Yu2017} or decoding strategies \citep{Schuez2021a}. 

Visual REG models usually process different forms of context information. Whereas some models encode differences in appearance between targets and surrounding objects \citep{Yu2016, Yu2017, Tanaka2019, Kim2020,Liu2020}, others use representations of the global image \citep{Mao2016, Luo2017, Zarriess2018, Panagiaris2020, Panagiaris2021}, both commonly supplemented with the relative position and size of the target in the image. 
On a conceptual level, however, recent work in visual REG generally follows the traditional paradigm by \citealt{Dale1995}, i.e. context is mainly considered in terms of so-called distractor or competitor objects, that are similar to the target and must therefore be excluded by naming differences (\citealt{Schuez2023}, but see \citealt{ilinykh-dobnik-2023-context} for context influences in object naming). 
In this view, context \enquote{exerts pressure}, as the speaker needs to reason about which attributes and words make the expression unambiguous \citep{CohnGordon2018a, Schuez2021a}. In this paper, we investigate how contextual information can be conceived as a resource that makes the generation of descriptions easier rather than harder.

\paragraph{Research gap} Little is known about how visual REG models internally exploit their context representations and in what way context exactly enhances the generation of expressions. 
A key difference to symbolic REG is that in visual REG failures in scene and object understanding due to e.g. imperfect visual input can lead to semantic errors, cf. \citet{Schuez2023}.
This is especially evident for the \textit{type} of objects: this attribute had a privileged role in early works \citep{Dale1995} as it is essential as the head of referential noun phrases. In visual REG, referents must first be correctly identified to \textit{name} them appropriately \citep{zarriess-schlangen-2017-obtaining, silberer-etal-2020-object, silberer-etal-2020-humans,ilinykh-dobnik-2023-context}, which is challenging in cases of deficient input, e.g. small or partially occluded objects \citep{Yao2010}.
In this paper, we aim to close this gap and investigate how visual context information helps REG models to be more resilient to deficits in their target inputs.

\section{Experimental Set-Up}
\label{sec:experimental_setup}
\subsection{Outline and Research Hypotheses}
  
    The main idea of this work is to train and test standard REG models on visual target representations occluded with varying amounts of noise, to investigate how different combinations of target and context can compensate for this perturbation. 
    For this, we draw on existing model architectures, and evaluate the trained models using both out-of-the-box quality metrics and more fine-grained human evaluation capturing the validity of assigned referent type labels, given the challenges of type identification in visual REG discussed in the previous section. 
    The evaluation results are also supported by supplementary analyses. 
    
    Generally, we expect that automatic metrics and human evaluation scores will drop for increasing amounts of target noise. However, we also hypothesize that visual context makes models more resilient, i.e., for the same amount of noise, 
    models supplied with context outperform variants with only target information. While we expect this general effect across all conditions, it should be more pronounced as the amount of occlusion increases.

\subsection{Models}

We set up two transformer-based REG models: 
TRF is a transformer model trained from scratch on REG data, CC builds upon a pre-trained language model. We define variants of both models using a) different combinations of target and context representations as the respective model inputs, and b) the amount of target noise during training and inference. Implementation and training details for our models can be found in appendix \ref{sec:model_appendix}.

Target representations include the visual contents of the target bounding box ($V_t$) and its location and size relative to the global image ($Loc_t$). As context representations, we use the embedding of the global image with the target masked out ($V_c$). 
We also experiment with symbolic representations about what kinds of objects the surrounding scene is composed of (\textit{scene summaries}, $S_c$). Incorporating symbolic scene features renders the task a multimodal fusion problem, i.e. the model has to align information from low-level visual and location information and symbolic scene summaries. 
Models processing only target information are indicated with the subscript $tgt$, whereas models processing $V_c$ and $S_c$ context information are indexed with $vis$ and $sym$, respectively.

To test our systems for perturbed target representations, we randomly replace a fixed proportion of the pixels in the bounding box with random noise during both training and inference. 
With this, we simulate cases of occlusion or other visual disturbances, which are common in real-world scenarios but rarely found in RefCOCO objects. 
We opted for pixel-wise occlusion for controllability reasons: Masking continuous sections would arguably be more akin to real-world occlusion by other objects, but could raise further questions, for example whether the parts masked out are important for determining the target class.
All systems are trained and tested with three noise settings: $0.0$ as our baseline setting, where no pixels are perturbed; $0.5$, where $50 \%$ of the pixels are replaced with noise; and $1.0$, where the entire content of the target bounding box is occluded, i.e. no visual target information is available,
similar in spirit to the \textit{Context-Obj} condition in \citet{ilinykh-dobnik-2023-context}. 
Importantly, models are trained separately for noise levels, i.e. a model evaluated for noise 0.5 is trained with the same noise level.

\paragraph{REG Transformer (TRF)}

We train a standard transformer architecture from scratch, which allows to carefully control and probe the effects of different target and context information.
We use the model from \citet{Schuez2023a}, which is based on an existing implementation for image captioning.\footnote{\href{https://github.com/saahiluppal/catr}{https://github.com/saahiluppal/catr}}
The model builds on ResNet \citep{He2015} encodings for targets and context, which are passed on to an encoder/decoder transformer in the style of \citet{Vaswani2017}, and is largely comparable to the system in \citet{Panagiaris2021}, but without self-critical sequence training and layer-wise connections between encoder and decoder. Unlike e.g. \citet{Mao2016}, we train the model using Cross Entropy Loss.

We compare three variants of this model, which take as input concatenated feature vectors comprised of the representations described above. TRF$_{tgt}$ receives only target information, i.e. an input vector $[V_t;Loc_t]$. TRF$_{vis}$ additionally receives visual context representations, namely $[V_t;Loc_t;V_c]$. TRF$_{sym}$ takes symbolic scene summaries as context, i.e. $[V_t;Loc_t;S_c]$. 

For both $V_t$ and $V_c$, the respective parts of the image are scaled to $224 \times 224$ resolution (keeping the original ratio and masking out the padding) and encoded with ResNet-152 \citep{He2015}, resulting in 196 features ($14 \times 14$) with hidden size $512$ for both target and context. 
$Loc_t$ is a vector of length 5 with the corner coordinates of the target bounding box and its area relative to the whole image, projected to the model's hidden size. 

The scene summary input for TRF$_{sym}$ consists of 134 features, representing the relative area each of the object or stuff categories in COCO occupies in the visual context. $S_c$ features are based on 2D panoptic segmentation maps (cf. Section \ref{sec:data}): We mask out the target bounding box and calculate the number of pixels assigned to each COCO category in the remaining image, then normalize the number of pixels assigned to each class by the total number of pixels. 
In TRF$_{sym}$, we add a further layer with jointly trained embeddings for all object and stuff types. 
In the model's forward pass, we concatenate all 134 embeddings, weighted by the respective coverage in the input image. 

\paragraph{Fine-tuned GPT-2 (CC)}

We adapt the \textit{ClipCap} model in \citet{Mokady2021} to the REG task. 
The authors use a simple MLP-based mapping network to construct fixed-size prefixes for GPT-2 \citep{Radford2019} from CLIP encodings \citep{Radford2021}, and fine-tune both the mapping network and the language model for the image captioning task.
To the best of our knowledge, this is the first model tested for REG which utilizes a pre-trained language model. 

As for the TRF model, we compare different variants of this base architecture. First, in CC$_{tgt}$, GPT-2 prefixes are constructed as $[V_t;Loc_t]$, where $V_t$ is computed like the CLIP prefix in the original paper (but for the contents of the target bounding box) and $Loc_t$ is the location features described above, projected into a single prefix token. In CC$_{vis}$, prefixes contain visual context representations, i.e. $[V_t;V_c;Loc_t]$. Here, $V_c$ is computed like $V_t$, but with a separate mapping network and with the global image (minus the target) as the visual input. Finally, CC$_{sym}$ includes symbolic scene summaries, i.e. $[V_t;S_c;Loc_t]$. Similar to the visual inputs, we use a mapping network to project the features before concatenation.

\subsection{Data}
\label{sec:data}
We use RefCOCO and RefCOCO+ \citep{Kazemzadeh2014} for training and evaluation. Both contain bounding boxes and expressions for the same objects in MSCOCO images \citep{Lin2014}, but while the location attributes \textit{left} and \textit{right} are highly frequent in RefCOCO, they have been excluded in RefCOCO+. 
The datasets contain separate \textit{testA} and \textit{testB} splits (1.9k and 1.8k items), where \textit{testA} only contains humans as referents and \textit{testB} all other object classes (but not humans).
To construct scene summaries ($S_c$) and analyze attention allocation patterns, we use annotations for panoptic segmentation \citep{Kirillov2018}, i.e. dense pixel-level segmentation masks for \textit{thing} and \textit{stuff} classes in MSCOCO images \citep{Caesar2016}. 

\subsection{Evaluation}
\label{sec:evaluation_criteria}

\paragraph{Generation Quality / N-Gram Metrics}
To estimate the general generation capabilities of our models we rely on BLEU \citep{papineni-etal-2002-bleu} and CIDEr \citep{Vedantam2014} as established metrics for automatic evaluation. 
As target occlusion involves random processes, we repeat inference ten times for all settings and average the results.

\paragraph{Referent Type Assignment / Human Evaluation}

To test whether our models succeed in assigning valid types to referents, we collect human judgments for generated expressions for a subset of 200 items from the RefCOCO \textit{testB} split, which is restricted to non-human referents.
Unlike for the automatic metrics, we use the results of a single inference run for each system.
The annotators were instructed to rate only those parts of the expressions that refer to the type of the referential target. For example, \enquote{the black dog} should be rated as correct if the target is of the type dog, but is actually white. 
All items should be assigned exactly one of the following categories:

  \begin{itemize}
    \item \textbf{Adequate / A}: The generated expression contains a valid type description for the referent.
    \item \textbf{Misaligned / M}: Type designators do not apply to the intended target, but to other objects (partially) captured by the bounding box.
    \item \textbf{Omission / O}: Omission of the target type, e.g. description via non-type attributes, pronominalization or general nouns such as \enquote{thing}.
    \item \textbf{False / F}: Type designations that do not apply to the intended target or other objects captured by the bounding box.
  \end{itemize}

Previous research has shown considerable variation in object naming (\citealt{silberer-etal-2020-object, silberer-etal-2020-humans}, among others). Therefore, for the \textit{A} category, type descriptions do not have to match the ground truth annotations, but different labels can be considered adequate if they represent valid descriptions of the target type. For example, \textit{dog}, \textit{pet} and \textit{animal} would be considered equally correct for depicted dogs.
Subsequent to the human evaluation, we investigate correlations between the evaluation results and further properties of the visual context.
  
\paragraph{Attention Allocation}

We also examine how our TRF$_{vis}$ model allocates attention over different parts of the input as a result of different noise levels during training. 
First, we follow \citet{Schuez2023a} in measuring the attention directed to the target and its context in both the encoder and decoder. 
For this, we compute $\alpha_t$, $\alpha_l$ and $\alpha_c$ as the cumulative attention weights directed to $V_t$, $Loc_t$ and $V_c$, respectively, normalized such that $\alpha_t + \alpha_l + \alpha_c = 1$.
We report the difference of attention directed to target and context, calculated as $\Delta_{t,c} = (\alpha_t + \alpha_l) - \alpha_c$, i.e. $0 < \Delta_{t,c} \leq 1$ if there is relative focus on the target, $-1 \leq \Delta_{t,c} < 0$ if there is relative focus on the context, and $\Delta_{t,c} = 0$ when both are weighted equally.
Second, we measure the model attention allocated to different classes of objects in the visual context, using the panoptic segmentation data described in Section \ref{sec:data}. 
Here, we first interpolate the model attention map to fit the original dimensions of the image and retrieve the respective segmentation masks. For each category $x \in X$, we then compute the cumulative attention weight $\alpha_{x}$ by computing the sum of pixels attributed to this category, weighted by the model attention scores over the image and normalized such that $\sum_{x \in X} \alpha_{x} = 1$. 
We report $\alpha_{x=tgt}$, i.e. attention allocated to areas of the visual context assigned \textit{the same category as the referential target}. 

\section{Results}

\subsection{Automatic Quality Metrics}
\label{sec:automatic_metrics}

\begin{table*}
  \centering
  \small
\begin{tabular}{lr|rrr|rrr|rrr|rrr}
\toprule
{} & {}    & \multicolumn{3}{|c|}{testA} & \multicolumn{3}{|c|}{testB} & \multicolumn{3}{|c|}{testA+} & \multicolumn{3}{c}{testB+} \\
{} & noise & Bl$_1$ & Bl$_2$ &   CDr & Bl$_1$ & Bl$_2$ &   CDr & Bl$_1$ & Bl$_2$ &   CDr & Bl$_1$ & Bl$_2$ &   CDr \\
\midrule \midrule 
TRF$_{tgt}$ &       &   0.55 &   0.35 &  0.86 &   0.57 &   0.35 &  1.28 &   0.49 &   0.31 &  0.77 &   0.36 &   0.19 &  0.68 \\
TRF$_{vis}$ &   0.0 &   0.58 &   0.39 &  0.93 &   0.61 &   0.39 &  1.36 &   0.50 &   0.32 &  0.83 &   0.37 &   0.20 &  0.73 \\
TRF$_{sym}$ &       &   0.54 &   0.34 &  0.84 &   0.57 &   0.35 &  1.27 &   0.46 &   0.29 &  0.78 &   0.37 &   0.19 &  0.72 \\
\midrule 
TRF$_{tgt}$ &       &   0.49 &   0.32 &  0.73 &   0.52 &   0.32 &  1.06 &   0.42 &   0.27 &  0.64 &   0.29 &   0.14 &  0.53 \\
TRF$_{vis}$ &   0.5 &   0.53 &   0.35 &  0.81 &   0.56 &   0.36 &  1.24 &   0.43 &   0.26 &  0.67 &   0.34 &   0.18 &  0.62 \\
TRF$_{sym}$ &       &   0.53 &   0.35 &  0.81 &   0.57 &   0.35 &  1.28 &   0.45 &   0.29 &  0.71 &   0.36 &   0.19 &  0.68 \\
\midrule 
TRF$_{tgt}$ &       &   0.35 &   0.17 &  0.34 &   0.30 &   0.14 &  0.20 &   0.29 &   0.15 &  0.20 &   0.07 &   0.01 &  0.04 \\
TRF$_{vis}$ &   1.0 &   0.46 &   0.29 &  0.60 &   0.55 &   0.36 &  1.14 &   0.32 &   0.17 &  0.34 &   0.29 &   0.14 &  0.47 \\
TRF$_{sym}$ &       &   0.42 &   0.24 &  0.51 &   0.53 &   0.33 &  1.12 &   0.31 &   0.15 &  0.31 &   0.30 &   0.14 &  0.48 \\
\midrule \midrule 
CC$_{tgt}$  &       &   0.48 &   0.30 &  0.70 &   0.47 &   0.28 &  0.88 &   0.42 &   0.27 &  0.70 &   0.29 &   0.14 &  0.53 \\
CC$_{vis}$  &   0.0 &   0.57 &   0.38 &  0.92 &   0.58 &   0.37 &  1.25 &   0.45 &   0.29 &  0.77 &   0.33 &   0.18 &  0.62 \\
CC$_{sym}$  &       &   0.45 &   0.28 &  0.66 &   0.56 &   0.36 &  1.22 &   0.44 &   0.28 &  0.73 &   0.37 &   0.20 &  0.70 \\
\midrule 
CC$_{tgt}$  &       &   0.38 &   0.21 &  0.48 &   0.36 &   0.20 &  0.51 &   0.40 &   0.25 &  0.64 &   0.27 &   0.14 &  0.47 \\
CC$_{vis}$  &   0.5 &   0.51 &   0.32 &  0.75 &   0.50 &   0.31 &  0.97 &   0.41 &   0.26 &  0.68 &   0.30 &   0.16 &  0.55 \\
CC$_{sym}$  &       &   0.44 &   0.27 &  0.61 &   0.57 &   0.36 &  1.17 &   0.35 &   0.21 &  0.46 &   0.33 &   0.17 &  0.57 \\
\midrule 
CC$_{tgt}$  &       &   0.35 &   0.16 &  0.37 &   0.29 &   0.12 &  0.16 &   0.27 &   0.14 &  0.20 &   0.10 &   0.02 &  0.06 \\
CC$_{vis}$  &   1.0 &   0.40 &   0.23 &  0.46 &   0.38 &   0.21 &  0.46 &   0.29 &   0.15 &  0.30 &   0.20 &   0.09 &  0.27 \\
CC$_{sym}$  &       &   0.42 &   0.25 &  0.52 &   0.55 &   0.34 &  1.17 &   0.31 &   0.16 &  0.32 &   0.32 &   0.16 &  0.53 \\
\bottomrule
\end{tabular}
  \caption{
  BLEU$_1$, BLEU$_2$ and CIDEr scores on RefCOCO testA and testB for all TRF and CC variants. 
  Systems indicated with \textit{tgt} can only access target information, \textit{vis} and \textit{sym} models are supplied with visual context and symbolic \textit{scene summaries}, respectively. Generally, context information leads to improved results, especially for high noise settings.}
  \label{tab:quality}
\end{table*}

Table \ref{tab:quality} shows the results of the automatic evaluation of our systems on the testA and testB splits in RefCOCO and RefCOCO+. 
Interestingly, the simpler TRF model outperforms CC, although the latter builds on pre-trained CLIP and GPT-2 which are known to be effective for image captioning \citep{Mokady2021}. Possible reasons for this can be seen in structural differences between bounding box contents and full images as used in the CLIP pre-training, or in higher compression when constructing the GPT prefixes.
Without target occlusion, model variants with access to visual context generally achieve the highest scores for both architectures (TRF$_{vis}$ and CC$_{vis}$, although CC$_{sym}$ exceeds the latter on testB+).

\begin{figure}
\centering
    \begin{subfigure}[b]{.9\columnwidth}
    \centering
    \includegraphics[width=\textwidth]{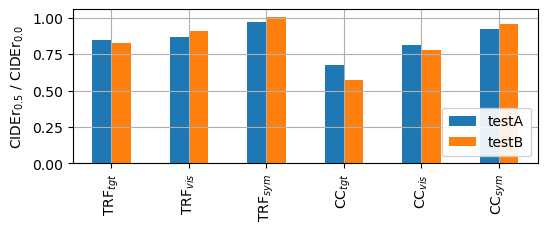}
    \caption{CIDEr for noise $0.5$, relative to noise $0.0$}
    \end{subfigure}

    \begin{subfigure}[b]{.9\columnwidth}
    \centering
    \includegraphics[width=\textwidth]{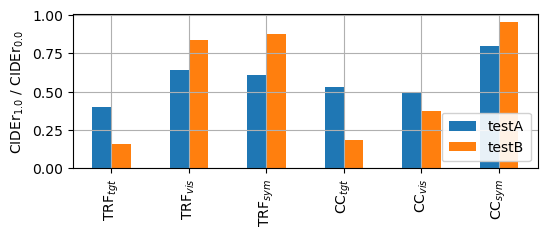}
    \caption{CIDEr for noise $1.0$, relative to noise $0.0$}
    \end{subfigure}
\caption{Relative CIDEr scores with respect to noise $0.0$ for RefCOCO testA and testB. For both TRF and CC, model variants with access to context are more robust against noise, especially for testB.}
\label{fig:cider_decline}
\end{figure}

As expected, scores consistently drop with increasing target noise. However, this is mitigated if context is available: For both TRF and CC, variants incorporating visual context are substantially more robust against target noise, even if target representations are entirely occluded, 
cf. Figure \ref{fig:cider_decline}. 
For example, for RefCOCO testB, CIDEr drops to $0.20$ for TRF$_{tgt}$ with noise 1.0 but TRF$_{vis}$ achieves scores as high as $1.14$, indicating that visual context combined with location features provides valuable information for describing (occluded) targets. 
Generally, TRF$_{vis}$ appears to be more effective at exploiting the visual context, e.g. CC$_{vis}$ with noise 1.0 drastically underperforms with CIDEr $0.46$ on testB. Although CC$_{tgt}$ is still outperformed (CIDEr $0.16$), this suggests problems for extracting relevant information from the visual context.

Similar patterns emerge when replacing visual context with symbolic \textit{scene summaries}: For both TRF and CC, model variants incorporating symbolic context features outperform their target-only counterparts in most cases, highlighting the potential of object co-occurrence information for making predictions robust to noise. For example, TRF$_{sym}$ achieves CIDEr $1.12$ for noise 1.0 in testB, comparable to TRF$_{vis}$. CC$_{sym}$ even outperforms CC$_{vis}$ for high noise settings (and all settings on testB+). On testB, CC$_{sym}$ scores are almost constant across noise levels, suggesting that the model is strongly relying on the scene summary information.

Interestingly, we see considerable differences between testA and testB: For both RefCOCO and RefCOCO+, target-only variants suffer less from occlusion on the testA splits (containing references to humans), but context is more effective on testB (containing references to other objects).
We hypothesize that models without meaningful visual input but access to location and size information can often \textit{guess right} on the frequent human classes in testA, but struggle with the higher variation in testB. Conversely, while human referents appear in a wide range of environments, other objects in testB rather tend to occur in specific surroundings, making context information more informative regarding their identity.

\subsection{Target Identification}
\label{sec:human_evaluation}

Human judgments were collected from 6 expert annotators, including the first author. Every system was evaluated independently by three annotators, with a Fleiss' Kappa of 0.85, indicating \textit{almost perfect} agreement \citep{Landis1977}. The final judgments are determined by majority vote.

The human evaluation results for the 200-item subset of RefCOCO testB are shown in Table \ref{tab:identification_scores}. 
Generally, we see similar patterns as in the BLEU and CIDEr scores discussed previously:
Ratios of \textit{\textbf{A}dequate} descriptions drop if noise ratios increase, while \textit{\textbf{F}alse} ratios increase at the same time. 
For \textit{\textbf{M}isalignments} and \textit{\textbf{O}missions}, higher noise generally leads to higher rates than the baseline setting. TRF$_{sym}$ and CC$_{sym}$ show particularly high \textit{M} rates for high noise settings, suggesting that the models often select object types that appear in the image, but not as the referent.
In the vast majority of cases, TRF variants outperform their CC counterparts. Again, the systems show large differences in exploiting visual context: Whereas CC$_{vis}$ assigns \textit{adequate} types in almost $20 \%$ of all cases for noise 1.0 (as compared to $0.5 \%$ without context information), TRF$_{vis}$ scores an impressive $66 \%$. 
 
Interestingly, symbolic scene summaries appear to be more effective for identification than visual context features: In most cases, models taking $S_c$ as input generate more adequate descriptions and fewer false descriptions and omissions than corresponding variants with visual context. For TRF$_{sym}$, this even extends to cases without target occlusion, unlike for BLEU and CIDEr (cf. Section \ref{sec:automatic_metrics}). Surprisingly, CC$_{sym}$ achieves very similar \textit{A} scores across all noise settings, narrowly exceeding TRF$_{sym}$ with noise 1.0. 
In line with the diminished influence of target occlusion observed for CIDEr and BLEU on testB, this indicates heavy reliance on symbolic scene representations (irrespective of the availability of visual target information), possibly due to problems with fusing symbolic (scene) and visual (target) information, a process that has received much attention in e.g. Visual Question Answering \citep{Zhang2019,Lu2023}.

\begin{table}
\centering
 \small
\begin{tabular}{lr|rrrr}
\toprule
{} &  noise &  \% A &  \% F &  \% O &  \% M \\
\midrule \midrule 
TRF$_{tgt}$ &        &     84.0 &      10.5 &     5.0 &        0.5 \\
TRF$_{vis}$ &    0.0 &     81.0 &      11.5 &     5.5 &        2.0 \\
TRF$_{sym}$ &        &     89.0 &       7.0 &     3.5 &        0.5 \\
\midrule 
TRF$_{tgt}$ &        &     66.5 &      28.0 &     4.0 &        1.5 \\
TRF$_{vis}$ &    0.5 &     70.5 &      18.5 &     7.0 &        4.0 \\
TRF$_{sym}$ &        &     81.5 &      14.5 &     2.5 &        1.5 \\
\midrule 
TRF$_{tgt}$ &        &      1.5 &      75.5 &    19.5 &        3.5 \\
TRF$_{vis}$ &    1.0 &     66.0 &      26.5 &     4.0 &        3.5 \\
TRF$_{sym}$ &        &     68.0 &      22.0 &     1.5 &        8.5 \\
\midrule \midrule 
CC$_{tgt}$  &        &     46.0 &      46.5 &     7.0 &        0.5 \\
CC$_{vis}$  &    0.0 &     75.5 &      21.5 &     3.0 &        0.0 \\
CC$_{sym}$  &        &     70.5 &      17.5 &     5.5 &        6.5 \\
\midrule 
CC$_{tgt}$  &        &     23.0 &      61.0 &    13.0 &        3.0 \\
CC$_{vis}$  &    0.5 &     55.5 &      35.5 &     6.5 &        2.5 \\
CC$_{sym}$  &        &     69.0 &      19.5 &     2.5 &        9.0 \\
\midrule 
CC$_{tgt}$  &        &      0.5 &      84.5 &    11.0 &        4.0 \\
CC$_{vis}$  &    1.0 &     19.5 &      68.5 &     9.0 &        3.0 \\
CC$_{sym}$  &        &     70.5 &      16.0 &     4.5 &        9.0 \\
\midrule \midrule 
$human$     &    0.0 &     90.5 &       2.5 &     6.0 &        1.0 \\
\bottomrule
\end{tabular}
 \caption{Ratios of \textbf{A}dequate, \textbf{F}alse, \textbf{O}mitted and \textbf{M}isaligned type descriptions (human annotation for 200 items from RefCOCO testB). Generally, contextual information leads to more adequate type descriptions, even if target representations are entirely occluded.}
 \label{tab:identification_scores}
\end{table}

\subsection{How do models exploit scene context?}
\label{sec:further_analyses}

So far, our results indicate that the scene context of referential targets greatly improves the resilience of REG models, to the extent that correct predictions are possible to a surprising rate even if target information is missing. Here, we aim to analyze how exactly contextual information is exploited by the models. 
As discussed in Section \ref{sec:background}, previous research indicates that regularities of object co-occurrence and scene properties facilitate e.g. object recognition in context. However, qualitative inspection indicates that for high noise, our systems often \textit{copy} from context, i.e. predict referent types that are also present in the surrounding scene, given that many classes of objects tend to appear in groups.
To investigate this, we (a) perform statistical tests to check whether similar objects in context support identification performance and (b) analyze the attention distribution for TRF$_{vis}$ to see how the respective context objects are weighted by the model.

\paragraph{Statistical analysis: Target categories in context}

We hypothesize that recalibration through context is more effective when the target class is also present in the scene. To test this, we conduct a correlation analysis between identification accuracy and the relative coverage of the target class in the context. For this, we again rely on panoptic segmentation annotations (cf. Section \ref{sec:data}) to compute the proportion of pixels of the same class as the referential target, normalized by the total size of the context. We binarize the human evaluation scores (\textit{True} if rated as \textit{A}, else \textit{False}) and compute the Point-biserial correlation coefficient between the relative coverage of the target class in context and the identification accuracy. The results are shown in Table \ref{tab:correlations}.
In almost all systems including visual or symbolic context representations, a higher prevalence of the target class in the visual context leads to significantly higher scores in human evaluation
($p < 0.05$ or higher significance for all systems except TRF$_{vis}$ / noise $0.0$ and CC$_{vis}$ / noise $1.0$), i.e. systems can easier compensate a lack of visual target information if the context contains similar objects. For TRF variants, the correlation is increasing with higher noise ratios, whereas it is more stable for CC. Interestingly, without access to context, both CC$_{tgt}$ and TRF$_{tgt}$ show weak correlation for the noise 0.0 setting (albeit only the former is significant), indicating the possibility of more general biases in the data.

\paragraph{Model attention to target category in context}

\begin{table}
  \centering
  \small
\begin{tabular}{lc|cc}
\toprule
{} &  noise &  corr. &  p  \\
\midrule \midrule 
TRF$_{tgt}$ &        &     0.128 &      --   \\
TRF$_{vis}$ &    0.0 &     0.109 &      --   \\
TRF$_{sym}$ &        &     0.154 &      $< 0.05$ \\
\midrule 
TRF$_{tgt}$ &        &     0.071 &      -- \\
TRF$_{vis}$ &    0.5 &     0.186 &      $< 0.01$ \\
TRF$_{sym}$ &        &     0.157 &      $< 0.05$ \\
\midrule 
TRF$_{tgt}$ &        &     0.046 &      -- \\
TRF$_{vis}$ &    1.0 &     0.321 &      $< 0.001$ \\
TRF$_{sym}$ &        &     0.277 &      $< 0.001$ \\
\midrule \midrule 
CC$_{tgt}$  &        &     0.156 &      $< 0.05$ \\
CC$_{vis}$  &    0.0 &     0.142 &      $< 0.05$ \\
CC$_{sym}$  &        &     0.353 &      $< 0.001$ \\
\midrule 
CC$_{tgt}$  &        &     0.049 &      -- \\
CC$_{vis}$  &    0.5 &     0.145 &      $< 0.05$ \\
CC$_{sym}$  &        &     0.249 &      $< 0.001$ \\
\midrule 
CC$_{tgt}$  &        &      0.045 &      -- \\
CC$_{vis}$  &    1.0 &      0.136 &      -- \\
CC$_{sym}$  &        &      0.246 &      $< 0.001$ \\
\bottomrule
\end{tabular}
  \caption{Correlation between identification accuracy and relative coverage of the target class in context. For most model variants with access to context, higher prevalence of the target class in the visual context leads to significantly higher scores in human evaluation.}
  \label{tab:correlations}
\end{table}

\begin{table}[]
    \centering
    \small
\begin{tabular}{lr|rr|rr}
\toprule
{} & {} & \multicolumn{2}{|c}{Encoder} & \multicolumn{2}{|c}{Decoder} \\
{} & noise & $\Delta_{t,c}$ & $\alpha_{x=tgt}$ & $\Delta_{t,c}$ & $\alpha_{x=tgt}$ \\
\midrule
$TRF_{vis}$ & 0.0 &           0.07 &            36.70 &           0.25 &            26.94 \\
$TRF_{vis}$ & 0.5 &          -0.30 &            35.27 &          -0.06 &            40.56 \\
$TRF_{vis}$ & 1.0 &          -0.17 &            35.63 &          -0.12 &            43.66 \\
\bottomrule
\end{tabular}
    \caption{
    Attention allocation scores for TRF$_{vis}$, averaged over RefCOCO testB. 
    $\Delta_{t,c}$ is the attention ratio between target and context, $\alpha_{x=tgt}$ is the \% of context attention directed to instances of the target class.
    }
\label{tab:attention_allocation}
\end{table}

In Table \ref{tab:attention_allocation}, we report the results of our attention analysis for TRF$_{vis}$ (cf. Section \ref{sec:evaluation_criteria}), averaged over all items in RefCOCO testB. For the target/context deltas $\Delta_{t,c}$, we expect that context is weighted more (i.e., scores are decreasing) as noise levels increase. Surprisingly, in the encoder, context is attended most in the $0.5$ noise setting. Decoder attention, however, follows our expected pattern.
Similarly, as shown by the $\alpha_{x=tgt}$ scores in Table \ref{tab:attention_allocation}, target noise does not seem to have a consistent effect on encoder attention to context objects sharing the target category. For the decoder, however, we see a notable increase: Whereas the baseline model assigns an average of 26.94 \% of its attention mass on context objects with the target class, this is significantly increased for higher noise settings (40.56 \% and 43.66 \%), suggesting that the TRF model learns to exploit the occurrence of similar objects in target and context as a common property of scenes in RefCOCO. 

\begin{figure*}
    \begin{subfigure}[b]{.30\textwidth}
        \centering
        \includegraphics[width=.8\columnwidth]{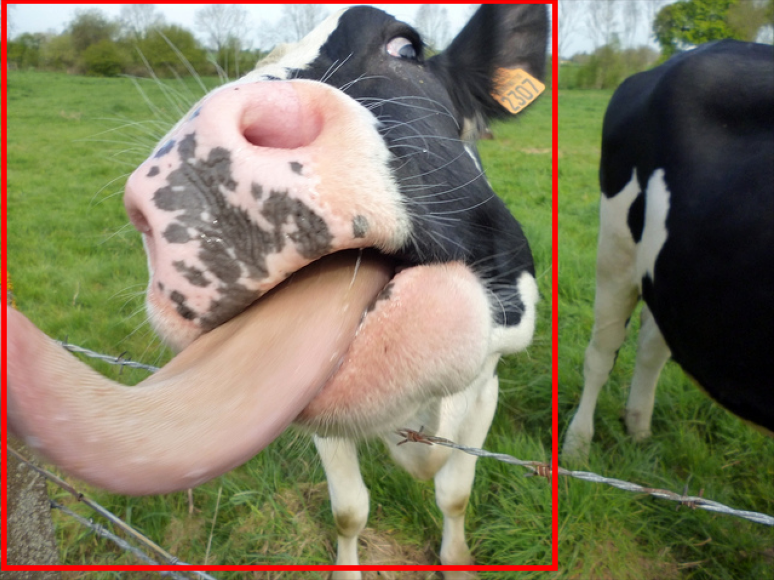}
        \addtolength{\tabcolsep}{-4pt}    
        \scriptsize
                \begin{tabular}{lllr}
                      & TRF$_{tgt}$ & cow (A)\\
            noise 0.0 & TRF$_{vis}$ & left cow (A)\\
                      & TRF$_{sym}$ & cow on left (A)\\
            \midrule
                      & TRF$_{tgt}$ & white horse (F)\\
            noise 0.5 & TRF$_{vis}$ & cow on left (A)\\
                      & TRF$_{sym}$ & cow (A)\\
            \midrule
                      & TRF$_{tgt}$ & man (F)\\
            noise 1.0 & TRF$_{vis}$ & left cow (A)\\
                      & TRF$_{sym}$ & cow on left (A)\\
        \end{tabular}
        \caption{Recognition errors for TRF$_{tgt}$ with target noise, mitigated by context.}
        \label{subfig:error1}
    \end{subfigure}
    \hspace{.03\textwidth}
    \begin{subfigure}[b]{.30\textwidth}
        \centering
        \includegraphics[width=.8\columnwidth]{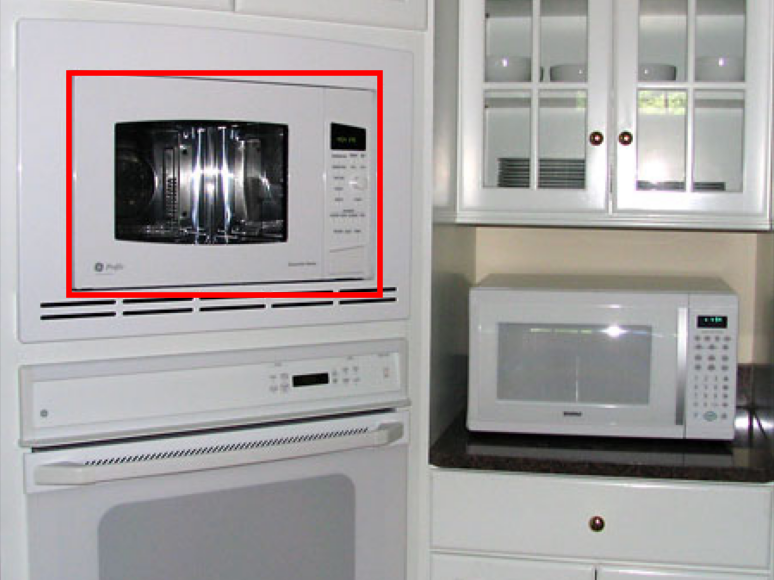}
        \addtolength{\tabcolsep}{-4pt}    
        \scriptsize
                \begin{tabular}{lllr}
                      & TRF$_{tgt}$ & top left micro (A)\\
            noise 0.0 & TRF$_{vis}$ & top left microwave (A)\\
                      & TRF$_{sym}$ & top left microwave (A)\\
            \midrule
                      & TRF$_{tgt}$ & left monitor (F)\\
            noise 0.5 & TRF$_{vis}$ & top microwave (A)\\
                      & TRF$_{sym}$ & top oven (F)\\
            \midrule
                      & TRF$_{tgt}$ & top left donut (F)\\
            noise 1.0 & TRF$_{vis}$ & top microwave (A)\\
                      & TRF$_{sym}$ & stove top (F)\\
        \end{tabular}
        \caption{TRF$_{sym}$ predictions are incorrect, but congruent with the scene.}
        \label{subfig:error2}
    \end{subfigure}
    \hspace{.03\textwidth}
    \begin{subfigure}[b]{.30\textwidth}
        \centering
        \includegraphics[width=.8\columnwidth]{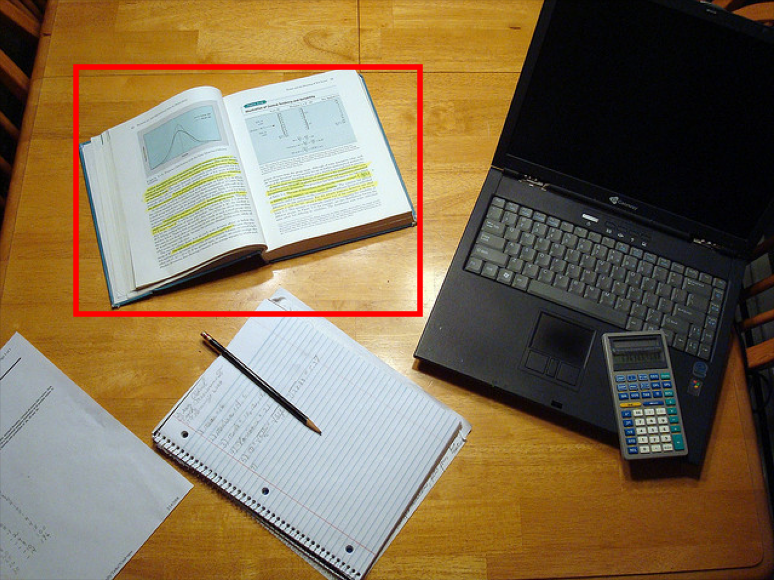}
        \addtolength{\tabcolsep}{-4pt}    
        \scriptsize
                \begin{tabular}{lllr}
                      & TRF$_{tgt}$ & top book (A)\\
            noise 0.0 & TRF$_{vis}$ & top book (A)\\
                      & TRF$_{sym}$ & paper on top (A)\\
            \midrule
                      & TRF$_{tgt}$ & white book (A)\\
            noise 0.5 & TRF$_{vis}$ & top laptop (F)\\
                      & TRF$_{sym}$ & open book (A)\\
            \midrule
                      & TRF$_{tgt}$ & top left (O)\\
            noise 1.0 & TRF$_{vis}$ & left laptop (F)\\
                      & TRF$_{sym}$ & laptop on left (F)\\
        \end{tabular}
        \caption{Copying errors (\textit{laptop}) for TRF$_{vis}$ and TRF$_{sym}$.}
        \label{subfig:error3}
    \end{subfigure}
    \caption{Examples from RefCOCO with generated expressions and human judgments (targets are marked red).}   
    \label{fig:errors}
\end{figure*}

\subsection{Qualitative Examples and Error Analysis}
\label{sec:error_analysis}

Figure \ref{fig:errors} shows expressions generated by all TRF variants and human identification judgments for three examples from RefCOCO.\footnote{For brevity, we present only expressions generated by TRF. For CC we observe similar patterns, the expressions can be found in Appendix \ref{app:cc_examples}.}
We identify both \textit{recognition errors}, where visual representations are incorrectly categorized, and \textit{inference errors}, where contextual information is misinterpreted.

Examples of recognition errors can be seen in Figure \ref{subfig:error1}, where TRF$_{tgt}$ predicts incorrect but visually related object types for noise 0.5 (\textit{horse}) and mostly unrelated types for noise 1.0 (\textit{man}). Here, both symbolic and visual context allow for robust predictions across noise levels. This is different in Example \ref{subfig:error2}: While similar problems can be seen for TRF$_{tgt}$ (\textit{monitor} instead of \textit{microwave} for noise 0.5), symbolic context leads to inference errors, i.e. TRF$_{sym}$ predicts incorrect object types that however fit into the general scene surrounding the target (\textit{oven} and \textit{stove top} as examples for kitchen appliances). 
Finally, in Example \ref{subfig:error3} we see evidence for the copying strategy discussed in Section \ref{sec:further_analyses}: With increasing noise, both TRF$_{vis}$ and TRF$_{sym}$ incorrectly predict \textit{laptop} as an object class present in the surrounding scene.

\section{Discussion and Conclusion}

Our findings show that contextual information about the surroundings of referents makes REG models more resilient against perturbations in visual target representations. 
Even if no target information is present at all, context allows REG models to maintain good results in automatic quality metrics and to identify referent types with high accuracy, as shown in the human evaluation results.
This holds for different kinds of context:
While especially the TRF$_{vis}$ model is able to leverage scene information from ResNet encodings of image contents outside the target bounding box, the same applies to symbolic scene representations, as included in TRF$_{sym}$ and CC$_{sym}$. 
This adds another perspective to basic assumptions of the REG paradigm, where context information is considered important mainly to ensure that references can be resolved without ambiguity. Here, we show, that it is also a valuable source for further communicative goals, i.e. the \textit{truthfulness} of generated expressions.

Interestingly, while related studies on human perception emphasize the importance of e.g. learned co-occurrence patterns between objects, our subsequent analysis rather points to implicitly learned copying strategies that appear to be highly effective for the relatively regular RefCOCO data. 
While this can also be seen as exploiting scene patterns, it is fundamentally different from the ways in which scene information is interpreted by humans (cf. Section \ref{sec:background}).
Therefore, we see an urgent need for data more representative of real-world scenarios to further investigate the impact of scene context on multimodal language generation. 

Overall, our results indicate that the influence of visual context in REG is more multifaceted than reflected in previous studies. 
Importantly, this study only provides an initial spotlight, as research in related fields suggests that there are other and more complex ways in which visual scene context may facilitate reference production.
With this in mind, we strongly advocate further research into scene context at the interface of perceptual psychology and V\&L generation.

\paragraph*{Risks and Ethical Considerations}

We do not believe that there are significant risks associated with this work, as we consider the generation of general expressions for generic objects in freely available datasets with limited scale. When selecting samples for human evaluation, we refrain from descriptions of people (that could potentially be perceived as hurtful). 
No ethics review was required. Our data does not contain any protected information and is fully anonymized. 

\paragraph*{Supplementary Materials Availability Statement:} 

\begin{itemize}
    \item RefCOCO and RefCOCO+ annotations and the RefCOCO API for computing BLEU and CIDEr scores are available on GitHub\footnote{\href{https://github.com/lichengunc/refer}{https://github.com/lichengunc/refer}}
    \item COCO images and panoptic segmentation annotations are available at \href{https://cocodataset.org/}{https://cocodataset.org/}
    \item Source code for the TRF base model are available on GitHub\footnote{\href{https://github.com/saahiluppal/catr}{https://github.com/saahiluppal/catr}}
    \item Source code for the CC base model are available on GitHub\footnote{\href{https://github.com/rmokady/CLIP_prefix_caption}{https://github.com/rmokady/CLIP\_prefix\_caption}}
    \item Our own code and data are available on GitHub\footnote{\href{https://github.com/clause-bielefeld/REG-Scene-Context}{https://github.com/clause-bielefeld/REG-Scene-Context}}
\end{itemize}

\section*{Acknowledgments}

This research has been funded by the Deutsche Forschungsgemeinschaft (DFG, German Research Foundation) – CRC-1646, project number 512393437, project B02.

\bibliography{document}

\appendix

\begin{table*}
    \small
\centering
\begin{tabular}{lc|cc|cc}
\toprule
{}             &  {}   & \multicolumn{2}{|c|}{RefCOCO} & \multicolumn{2}{|c}{RefCOCO+}  \\
{}             &  noise& epochs    & CIDEr (val)       &    epochs      & CIDEr (val)    \\
\midrule
TRF$_{tgt}$    &   0.0 &       8   &   1.074           &    7           &       0.803    \\
TRF$_{vis}$    &   0.0 &       6   &   1.156           &    7           &       0.828    \\
TRF$_{sym}$    &   0.0 &       8   &   1.075           &    5           &       0.794    \\
\midrule
TRF$_{tgt}$    &   0.5 &       11  &   0.936           &    4           &       0.647    \\
TRF$_{vis}$    &   0.5 &       9   &   1.035           &    11          &       0.697    \\
TRF$_{sym}$    &   0.5 &       14  &   1.032           &    10          &       0.74     \\
\midrule
TRF$_{tgt}$    &   1.0 &       5   &   0.302           &    3           &       0.173    \\
TRF$_{vis}$    &   1.0 &       6   &   0.869           &    5           &       0.449    \\
TRF$_{sym}$    &   1.0 &       12  &   0.818           &    5           &       0.45     \\
\midrule \midrule
CG$_{tgt}$     &   0.0 &       7   &   0.824           &    4           &       0.673    \\
CG$_{vis}$     &   0.0 &       4   &   1.103           &    5           &       0.754    \\
CG$_{sym}$     &   0.0 &       8   &   0.908           &    8           &       0.756    \\
\midrule
CG$_{tgt}$     &   0.5 &       8   &   0.554           &    14          &       0.603    \\
CG$_{vis}$     &   0.5 &       10  &   0.894           &    5           &       0.679    \\
CG$_{sym}$     &   0.5 &       11  &   0.89            &    11          &       0.553    \\
\midrule
CG$_{tgt}$     &   1.0 &       2   &   0.294           &    4           &       0.174    \\
CG$_{vis}$     &   1.0 &       7   &   0.526           &    11          &       0.334    \\
CG$_{sym}$     &   1.0 &       9   &   0.823           &    8           &       0.45     \\
\bottomrule
\end{tabular}
    \caption{Training information for all TRF and CC variants. CIDEr scores are computed for the val splits in RefCOCO / RefCOCO+.}
    \label{tab:model_training}
\end{table*}

\section{Limitations}

We identify the following limitations in our study: 

First, in both training and evaluation, we do not consider pragmatic informativeness as a core criterion for the REG task. We train our models using Cross Entropy Loss and do not test whether the generated expressions unambiguously describe the referential target, instead focusing on semantic adequacy as an important prerequisite for the generation of successful referential expressions. However, we acknowledge that a comprehensive view would require the consideration of both semantic and pragmatic aspects. 

Also, we do not consider recent developments such as multimodal LLMs, although the high diversity of their training data would contribute an interesting aspect to this study. Here, we selected our models with a focus on both modifiability and transparent processing.

Finally, additional vision and language datasets such as VisualGenome \citep{Krishna2016} would have made the results more representative. However, due to time and space constraints, we leave this for future research.

\section{Model implementation and training}
\label{sec:model_appendix}

For the hyperparameters of our models, we largely followed \citet{Panagiaris2021} (TRF) and \citet{Mokady2021} (CC). During inference, we relied on greedy decoding.

The TRF model has 3 encoder and 3 decoder layers with 8 attention heads, hidden dimension and feedforward dimension of 512, and was trained with an initial learning rate of 0.0001 for the transformer encoder and decoder, and 0.00001 for the pre-trained ResNet-152 backbone. Our TRF models have approximately 103,000,000 parameters.

For our CC model, we kept the settings defined by \citet{Mokady2021}. From the two models proposed in this work, we used the variant where a simple MLP is used as a mapping network and the GPT-2 language model is fine-tuned during training. However, we have different prefix sizes than in the original paper: For CC$_{tgt}$, we have a prefix size of 11, i.e. 10 for the visual target representation and 1 for the target location information. For CC$_{vis}$ and CC$_{sym}$, our prefix size is 21, with additional 10 tokens for the context. The model was trained using a learning rate of 0.00001. CC$_{vis}$ has approximately 338,000,000, CC$_{sym}$ has 337,000,000 and CC$_{tgt}$ has 307,000,000 parameters.

We trained our models on an Nvidia RTX A40. Both RefCOCO and RefCOCO+ contain approximately 42k items for training. The number of training epochs per system and the final CIDEr scores over the validation sets are displayed in Table \ref{tab:model_training}. We trained all our models for a maximum of 15 epochs, with early stopping if no new maximum for CIDEr over the validation set has been achieved for three consecutive epochs. Per epoch, the compute time was approximately 2.30 h for all systems.

\begin{figure*}
    \begin{subfigure}[b]{.30\textwidth}
        \centering
        \includegraphics[width=.8\columnwidth]{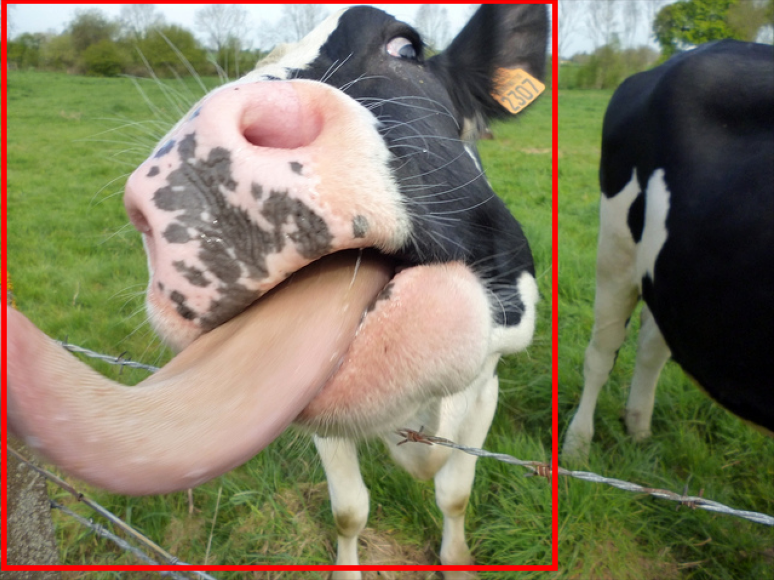}
        \addtolength{\tabcolsep}{-4pt}    
        \scriptsize
        \begin{tabular}{lllr}
              & CC$_{tgt}$ & left bird (F)\\
    noise 0.0 & CC$_{vis}$ & white cow (A)\\
              & CC$_{sym}$ & cow on left (A)\\
    \midrule
              & CC$_{tgt}$ & left giraffe (F)\\
    noise 0.5 & CC$_{vis}$ & left cow (A)\\
              & CC$_{sym}$ & cow on left (A)\\
    \midrule
              & CC$_{tgt}$ & left guy (F)\\
    noise 1.0 & CC$_{vis}$ & cow on left (A)\\
              & CC$_{sym}$ & cow on left (A)\\
\end{tabular}
    \end{subfigure}
    \hspace{.03\textwidth}
    \begin{subfigure}[b]{.30\textwidth}
        \centering
        \includegraphics[width=.8\columnwidth]{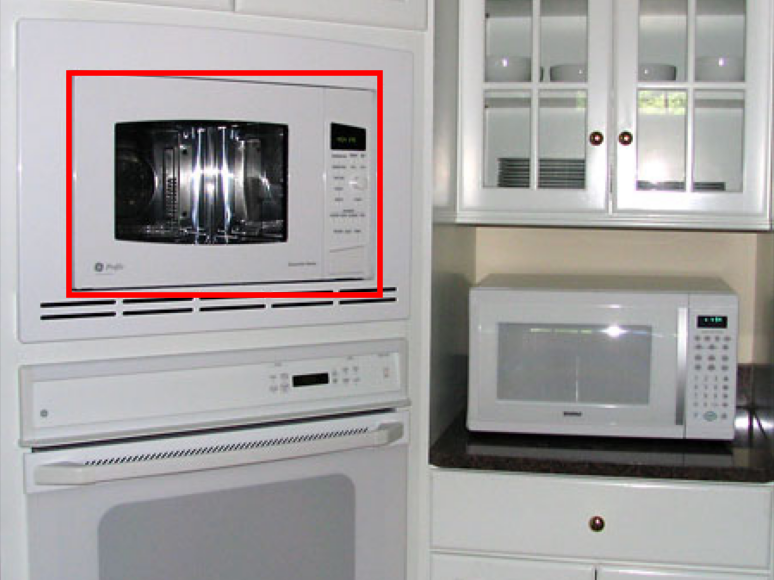}
        \addtolength{\tabcolsep}{-4pt}    
        \scriptsize
        \begin{tabular}{lllr}
          & CC$_{tgt}$ & left one (O)\\
noise 0.0 & CC$_{vis}$ & top microwave (A)\\
          & CC$_{sym}$ & left stove (F)\\
\midrule
          & CC$_{tgt}$ & left clock (F)\\
noise 0.5 & CC$_{vis}$ & left microwave (A)\\
          & CC$_{sym}$ & stove top (F)\\
\midrule
          & CC$_{tgt}$ & top left donut (F)\\
noise 1.0 & CC$_{vis}$ & left umbrella (F)\\
          & CC$_{sym}$ & top left stove (F)\\
\end{tabular}
    \end{subfigure}
    \hspace{.03\textwidth}
    \begin{subfigure}[b]{.30\textwidth}
        \centering
        \includegraphics[width=.8\columnwidth]{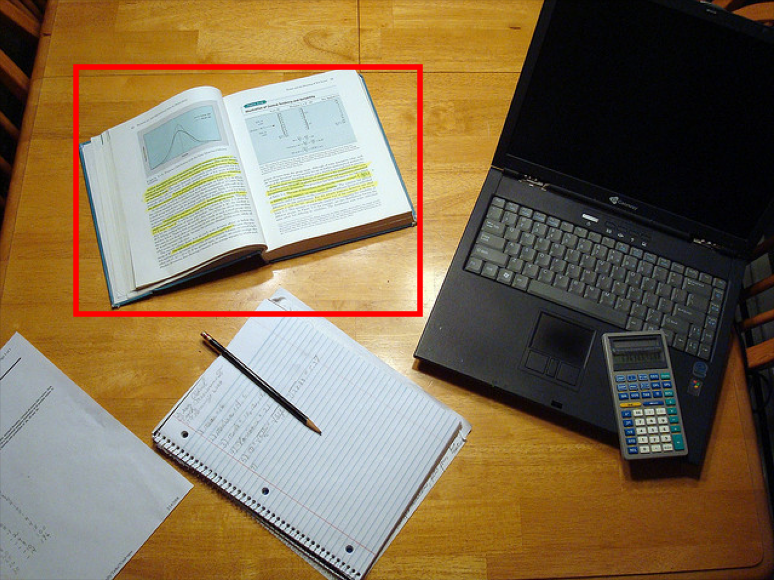}
        \addtolength{\tabcolsep}{-4pt}    
        \scriptsize
        \begin{tabular}{lllr}
              & CC$_{tgt}$ & left monitor (F)\\
    noise 0.0 & CC$_{vis}$ & book on left (A)\\
              & CC$_{sym}$ & left laptop (F)\\
    \midrule
              & CC$_{tgt}$ & left monitor (F)\\
    noise 0.5 & CC$_{vis}$ & keyboard on left (F)\\
              & CC$_{sym}$ & left laptop (F)\\
    \midrule
              & CC$_{tgt}$ & top left donut (F)\\
    noise 1.0 & CC$_{vis}$ & left laptop (F)\\
              & CC$_{sym}$ & left laptop (F)\\
\end{tabular}
    \end{subfigure}
    \caption{Examples from RefCOCO with expressions generated by CC variants and human judgments (targets are marked red).}   
    \label{fig:errors_cc}
\end{figure*}

\section{Scientific Artifacts}

In our work, we mainly used scientific artifacts in the form of existing model implementations, all of which are cited or referenced in Section \ref{sec:experimental_setup}. The model implementations were published under permissive licences, i.e. \textit{MIT} (TRF) and \textit{Apache 2.0} (CC). We publish our modifications to the model implementations using the same licences, and our other code and data using permissive licences.

Apart from this, we relied on scikit-learn (version 1.2.0, \citealt{Pedregosa2011}) for our statistic analysis and the RefCOCO API \citep{Kazemzadeh2014, Yu2016}\footnote{https://github.com/lichengunc/refer} for computing BLEU and CIDEr scores.

\section{Human Evaluation}

We conducted a human evaluation in which the adequacy of assigned referent types in English referring expressions was assessed. The annotation guidelines are published in our code repository.

Our annotators were undergrad student assistants from linguistics and computational linguistics, which were paid by the hour according to the applicable pay scale. The annotators were informed about the intended use of their produced data. Along with our code, we publish the fully anonymized raw and aggregated results of the human evaluation. 

\section{Qualitative Examples for CC}
\label{app:cc_examples}

In Section \ref{sec:error_analysis} we presented expressions generated by all TRF variants and discussed different types of errors in the model outputs. CC responses for the same examples are shown in Figure \ref{fig:errors_cc}. In general, we observe similar patterns as for TRF, but with some additional errors (especially for CC$_{tgt}$). 

\end{document}